\documentclass{article}

\usepackage[letterpaper, left=1in, right=1in, top=1in, bottom=1in]{geometry}
\usepackage{authblk} 
\usepackage{soul}\setuldepth{article}
\usepackage{hyperref}
\usepackage{pdfcomment}
\usepackage[color=orange!40]{changes}

\usepackage{graphicx} 

\title{Deep learning and abstractive summarisation for radiological reports: an empirical study for adapting the PEGASUS models' family with scarce data.}
\date{\today}  
\author[1]{Claudio Benzoni\thanks{\textbf{Corresponding Author} \\ \textbf{Claudio Benzoni, MSc, PhD} \\ Institute of AI and Informatics in Medicine (AIIM) \\
TUM University Hospital, Technical University of Munich, Munich, Germany \\
Grillparzerstr. 18 / 3. OG \\
81675 Munich \\
Tel.: +49 89 4140 4322\\
Email: claudio.benzoni@tum.de \\
ORCID iD: 0009-0002-0929-540X}}

\author[2]{Martina Langhals}
\author[1]{Martin Boeker} 
\author[1]{Luise Modersohn}
\author[2]{Máté E. Maros}

\affil[1]{Chair of Medical Informatics, Institute of Artificial Intelligence and Informatics in Medicine (AIIM), TUM University Hospital, Technical University of Munich, Munich, Germany}
\affil[2]{Department of Biomedical Informatics, Research Group MIDorAI, Medical Faculty Mannheim, Heidelberg University, Mannheim, Germany}

\begin{document}

\maketitle

\noindent \textbf{Short title}: PEGASUS models and overfitting
\noindent \textbf{Word count}: 3194 \\
\noindent \textbf{Numbers of figures \& tables}: 4 \& 3 \\
\noindent \textbf{Note}: The authors produced this text without using generative AI.
\section{Abstract}
Regardless of the rapid development of artificial intelligence, abstractive summarisation is still challenging for sensitive and data-restrictive domains like medicine \cite{van_veen_adapted_2024}. With the increasing number of imaging, the relevance of automated tools for complex medical text summarisation is expected to become highly relevant. In this paper, we investigated the adaptation via fine-tuning process  of a non-domain-specific abstractive summarisation encoder-decoder model family, and gave insights to practitioners on how to avoid over- and underfitting. We used PEGASUS \cite{Zhang2019PEGASUSPW} and PEGASUS-X \cite{phang2022investigating}, on a medium-sized radiological reports public dataset. For each model, we comprehensively evaluated two different checkpoints with varying sizes of the same training data. We monitored the models' performance with lexical and semantic metrics during the training history on the fixed-size validation set. PEGASUS exhibited different phases, which can be related to epoch-wise double-descent, or peak-drop-recovery behaviour. For PEGASUS-X, we found that using a larger checkpoint led to a performance detriment. This work highlights the challenges and risks of fine-tuning models with high expressivity when dealing with scarce training data, and lays the groundwork for future investigations into more robust fine-tuning strategies for summarisation models in specialised domains.

\section{Introduction}

As medical imaging becomes increasingly complex and voluminous, there is a growing need for robust, automated solutions capable of distilling clinically relevant information from intricate textual data \cite{nishio2024fully}.

Classical natural language processing (NLP) \cite{schutze2008introduction} and traditional machine learning (ML) techniques have long been applied to medical text in the field of radiological reports \cite{pons2016natural} \cite{maros2021comparative}. However, in recent years, the field has greatly benefited from adopting neural networks and deep learning (DL) techniques. In fact, for the first time in 2017, it was shown \cite{doi:10.1148/radiol.2017171115} that a convolutional neural network (CNN) was superior in classifying the presence of pulmonary embolism with respect to the back-then state-of-the-art PeFinder \cite{CHAPMAN2011728}. While the latter was based on more traditional NLP approaches, e.g. grammatical feature definitions, concept codes and others, the CNN required only binary labels and it could learn the essential patterns in classification via training. 
A significant leap for DL for text-related tasks was the advent of the encoder-decoder \textit{transformer} architecture \cite{vaswani2017attention}. This advancement sparked the development of massive pre-trained models that either use only the encoder (BERT \cite{2018arXiv181004805D}, modernBERT \cite{warner2024smarterbetterfasterlonger}, \cite{ModernBERT}), or only the decoder (GPTs \cite{radford2019language}, \cite{2020arXiv200514165B}, \cite{openai2024gpt4technicalreport}) part of the original transformer\footnote{Recently, it became popular to refer to decoder-only architectures as Large Language Models, despite the original architecture surely fitting the definition \cite{decoder-only}.}. Nowadays, these models have successfully replaced the CNN (and Long Short-Term Memory \cite{10.1162/neco.1997.9.8.1735}) architecture  in the general text as well as in radiological reports domains \cite{chaves2023rales}, \cite{van_veen_adapted_2024}. Although transformer-based decoder-only language models (LLMs) have advanced considerably, generating abstractive summaries in high-stakes, data-sensitive fields like medicine remains challenging and requires additional domain-specific knowledge. This can be injected either with model-specific prompting and in-context learning, or with full fine-tuning (FT). In both cases, the sheer size of LLMs requires dedicated hardware (high-performance GPUs) that is not generally available in the clinical setting. 
On the contrary, the information bottleneck of the encoder-decoder architecture is particularly efficient and tailored for \textit{seq2seq} tasks as summarisation and machine translation \cite{jurafsky2025speech}.

Therefore, in this paper, we used PEGASUS and PEGASUS-X, two encoder-decoder models pretrained for abstractive text summarisation \cite{Zhang2019PEGASUSPW} \cite{phang2022investigating}, the latter being an extension of the first capable of handling up to 32 times more input tokens. We adapted the models via FT and evaluated their summarisation performance on a publicly available dataset of X-ray reports.

Our work builds upon a previous study \cite{dai-etal-2021-bdkg}, which has partially fine-tuned the PEGASUS model on the same public dataset, and assessed the quality of the summarisation, using only the ROUGE score \cite{MEDIQA}. We substantially extended the scope of the existing work by complementing it with more metrics, showing the pitfalls of naive fine-tuning, and suggesting that a larger checkpoint does not necessarily lead to improvements.

The objective of this work is to highlight the strengths of the computationally efficient encoder-decoder architecture and specialised models for summarisation tasks, to provide valuable insights for practitioners  about under- and overfitting phenomena when performing domain adaptation on limited training data, and to suggest the monitoring of the training history as a viable way to prevent them.

\section{Methods}
\subsection{Reference cohort}\label{reference_cohort}
We used the publicly available anonymised posterior-anterior chest X-ray studies collected in \cite{jaeger2013automatic} and publicly released in \cite{open-i}. This dataset comprises 3996 radiology reports and 8121 related radiological images of various patients collected from two hospitals in the Indiana Network for Patient Care Database. We downloaded the preprocessed text data for \cite{van_veen_adapted_2024} and available at the project's repository \cite{van_veen_adapted_2024_github}, which reduced the original dataset to  85\% of the original cohort. We kept the identical train/validation/test splits of 80/10/10 to ensure comparability. While we did not perform hyperparameter tuning, we used the validation split to assess the models' performance during the fine-tuning process  \cite{kocak2023checklist}. 

\begin{figure}[htbp!]
  \centering
  \begin{minipage}[b]{0.48\textwidth}
    \centering
    \includegraphics[width=\textwidth]{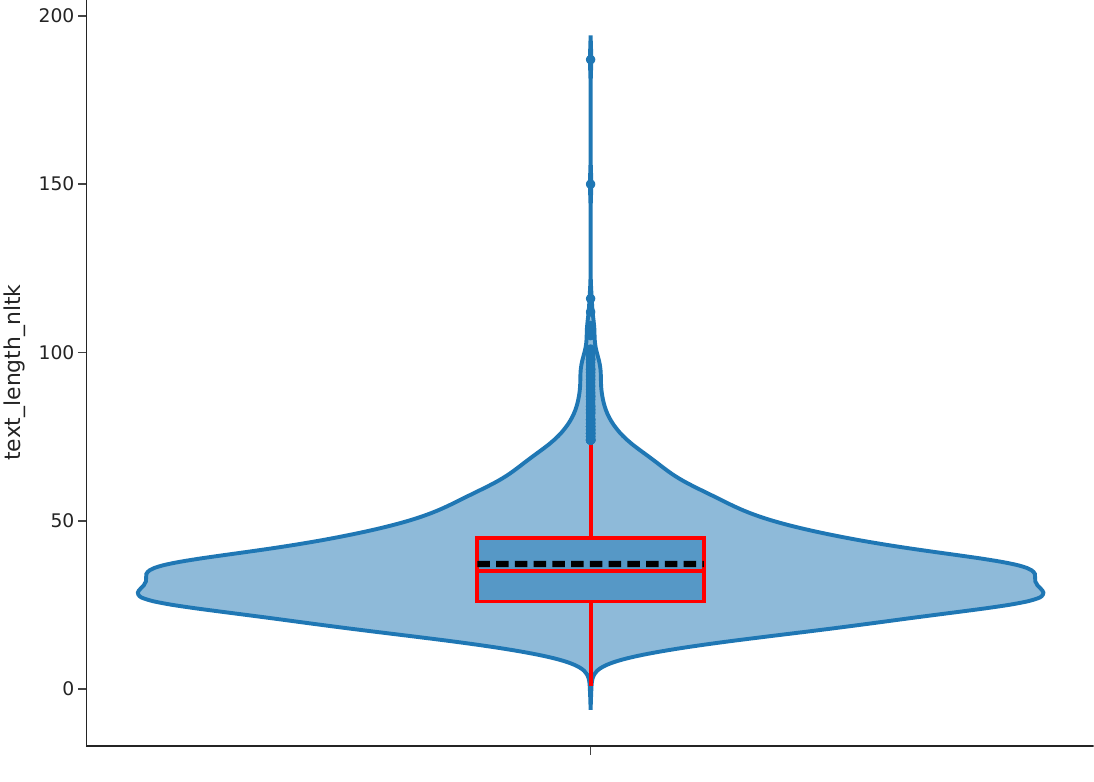}
    \caption{Texts‐lengths distribution.}
    \label{fig:violin_text}
  \end{minipage}
  \hfill
  \begin{minipage}[b]{0.48\textwidth}
    \centering
    \includegraphics[width=\textwidth]{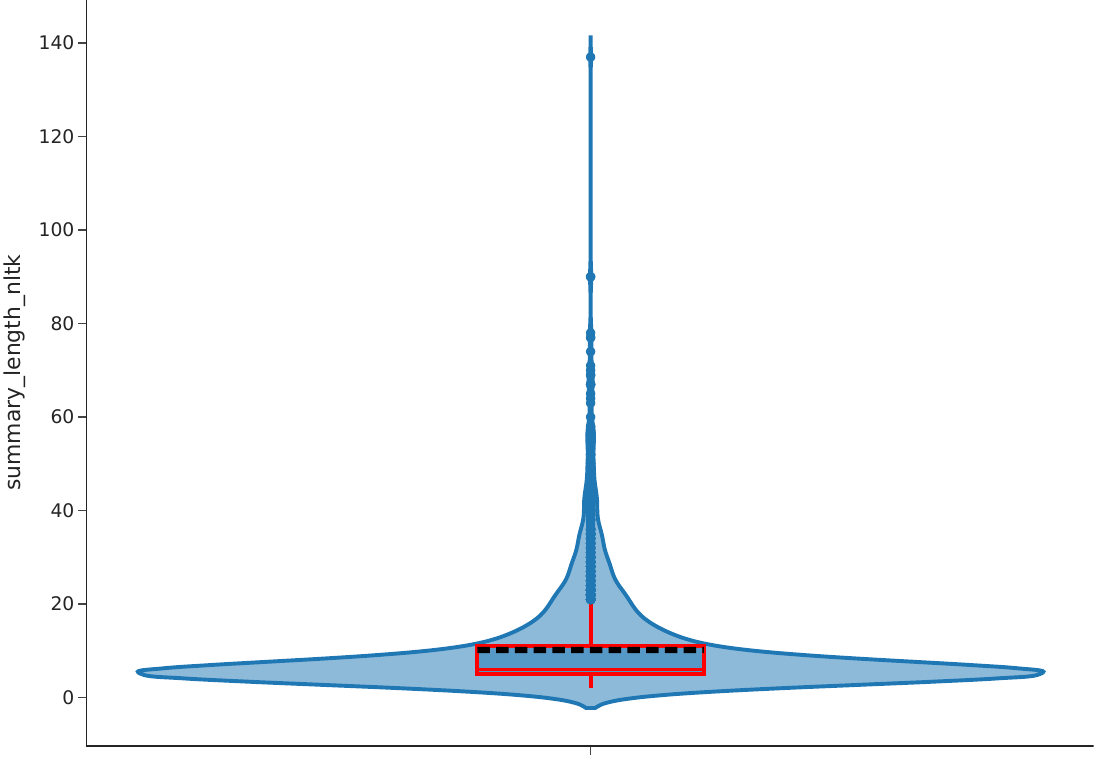}
    \caption{Summary‐lengths distribution.}
    \label{fig:violin_summary}
  \end{minipage}
  \caption{Side‐by‐side distributions of text lengths (left) vs.\ summary lengths (right) in the train split. 
  In solid red: box plots enclose the range (Q1, Q3), while whiskers extend to the last point within 1.5 IQR from the box's margins. In soft indigo: violin plots created with a Gaussian kernel; the respective KDE bandwidth is according to Silverman’s rule (left: 3.32, right: 2.12). Single outliers are drawn beyond the whiskers. In dashed black: the mean.}
  \label{fig:violins_comparison}
\end{figure}

In this case, the summarisation process accounts for proceeding from longer texts (findings) to shorter ones (impressions) while retaining the semantic meaning of the former. We manually identified three erroneous summaries that were longer than the original text, and we excluded them. 
We studied the distribution of lengths for the tokenized\footnote{Tokenization consists of separating sentences into smaller pieces.} versions of texts and summaries in the train split using the well-established word tokenizer \textit{nltk} \cite{bird2009natural}. As we can see in Fig.  \ref{fig:violins_comparison} both distributions are skewed, with the median for the text being 35 tokens, and the median for summaries being 6 tokens. 

Summaries have a more peaked distribution because of the negated diagnoses, which are generally short. We searched through the most popular 125 categories of summary lengths and manually labelled the negated  cardiopulmonary diagnoses, i.e. \textit{"no acute cardiopulmonary abnormality"}, \textit{"no active disease"}, \textit{"no acute radiographic cardiopulmonary process"}, and possible synonyms and/or punctuation variations. This list is not exhaustive, but it allows a better understanding of the dataset. According to this binary classification, in the train split, 63.3\% are negated diagnoses, 36.7\% are diagnoses. The validation split has a slightly different distribution: negated diagnoses are 61.9\%, diagnoses are 38\%.  

\subsection{Selection of data subsets}
\begin{figure}[h!]
    \centering
    \includegraphics[width=0.45\linewidth]{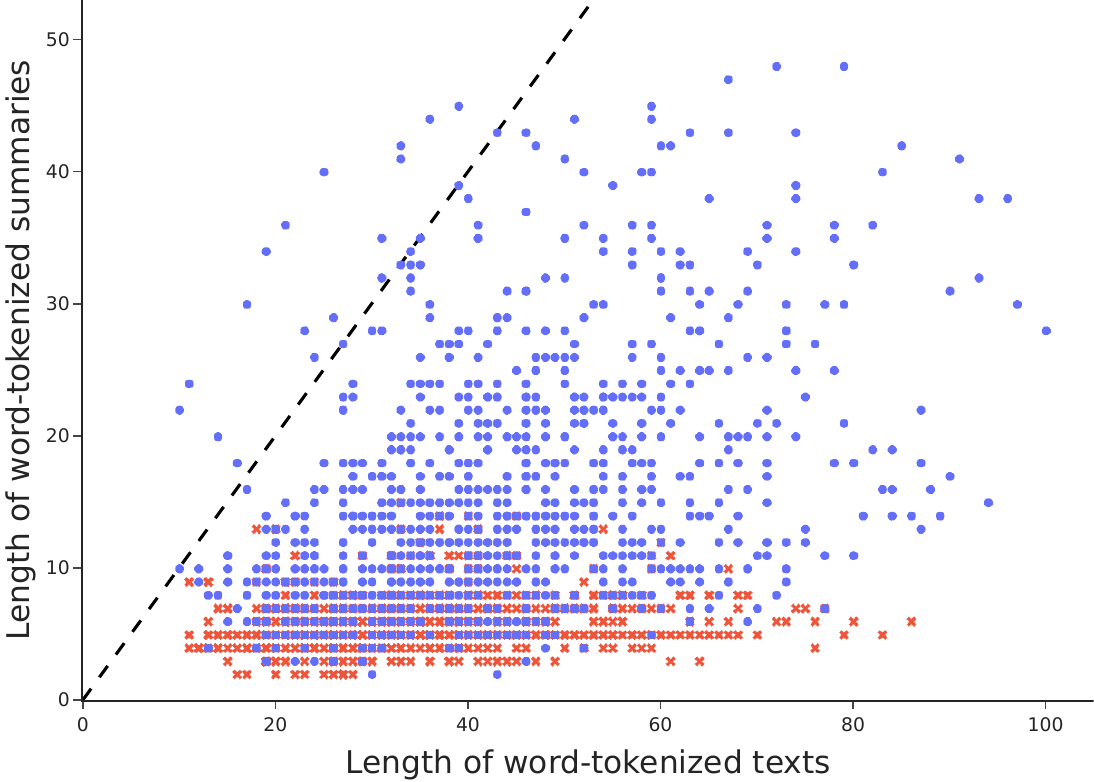}
    \caption{98\% percentile of the preprocessed training split. Most of the data points lie under the equal-length line. Negated diagnoses are displayed in red crosses, diagnoses in blue circles.}
    \label{fig:preprocessed_dataset}
\end{figure}

Encoder-decoder models require that all the inputs in a batch must be of the same size \cite{huggingfaceConcept}. To achieve this, a consistent maximum length for the truncation and padding of the transformer-related tokenizer has to be chosen. On the one hand, the truncation and padding lengths must be large enough to avoid cutting part of the text to process. On the other hand, since attention in transformers scales quadratically with the tokenized sequence length \cite{vaswani2017attention}, the maximum length should be as small as possible to minimise the computational requirements. To prevent outliers from dictating this computational budget choice,  we used a standard multivariate anomaly detection approach based on the Mahalanobis distance \cite{MLPlusMahalanobis}, \cite{10.1002/widm.1236} with regard to the word-tokenized text and summary lengths. We selected the data points in the 98\% percentile, see Fig. \ref{fig:preprocessed_dataset}. 

\subsection{Models}
We used two models, and two checkpoints per model. For PEGASUS, we employed \textit{pegasus-large} and \textit{pegasus-xsum} \cite{Zhang2019PEGASUSPW}. The second checkpoint was trained on the XSum dataset \cite{narayan-etal-2018-dont} and then reflects its characteristics, favouring the creation of highly concise one-line summaries. For PEGASUS-X we used \textit{pegasus-x-large} and \textit{pegasus-x-base}. Apart from the last one, which has 272 million parameters, all the other checkpoints contain 568 million parameters. 
These models' checkpoint sizes make them suitable for training all the architecture's weights. We believe this procedure is recommendable whenever it is necessary to adapt a task-specific model to a highly specialised domain \cite{gururangan2020dontstoppretrainingadapt}.

PEGASUS and PEGASUS-X use SentencePiece \cite{kudo-richardson-2018-sentencepiece}, which implements two subword tokenization algorithms including byte-pair encoding \cite{sennrich-etal-2016-neural} and unigram-language model \cite{DBLP:journals/corr/abs-1804-10959}. As we expect that subword tokenization creates more tokens than \textit{nltk}, we accordingly chose as truncation and padding lengths 1.33 times the respective word-tokenized maximum, a common rule-of-thumb ratio \cite{075rule-of-thumb}.

We fine-tuned the above checkpoints for 300 epochs with a nested family of subsets 
representing 10\%, 50\% and 100\% of the original training split\footnote{To ensure reproducibility, we hereby report the hyperparameters we used: \textit{learning\_rate}=2e-5, \textit{fp16}=True,  \textit{weight\_decay}=0.01, \textit{predict\_with\_generate}=True, \textit{generation\_max\_length}=64, \textit{generation\_num\_beams}=4, and
\begin{itemize}
\item pegasus-large/pegasus-xsum: \textit{per\_device\_train\_batch\_size}=34, \textit{per\_device\_eval\_batch\_size}=4,
\item pegasus-x-base: 
\textit{per\_device\_train\_batch\_size}=26, \textit{per\_device\_eval\_batch\_size}=2, 
\item pegasus-x-large:
\textit{per\_device\_train\_batch\_size}=10, 
\textit{per\_device\_eval\_batch\_size}=2.
\end{itemize}} We evaluated the metrics of interest on the validation dataset at every epoch to assess the possible effects of under- and overfitting. We saved the models and the trainers' states every five epochs. 
On our infrastructure (a cluster of four A40 GPUs), the full fine-tuning took less than one day. For fine-tuning, inference, and evaluation, we used the \textit{Huggingface} library.
\subsection{Evaluation for abstractive summarisation}
Evaluation for Natural Language Generation is an astoundingly changing field of research on its own \cite{10.1145/3485766}. Here, we focused on traditional metrics that compare models' predictions to labels. The labels we considered were the impressions written by the original radiologist during their daily practice, considered as the gold standard. We can divide the evaluation metrics we employed into two main classes as follows.

Metrics that are based on the N-gram overlap between the target and references represent the standard starting point of evaluation benchmarks. BLEU is based on precision and contains a brevity penalty \cite{papineni-etal-2002-bleu}. Due to its high sensitivity to tokenization, we used its standardised implementation SacreBLEU \cite{post-2018-call}. ROUGE is similar to BLEU, being based on N-gram overlaps, but it is focused on \textit{recall}. Hence, ROUGE is the common choice for summarisation tasks \cite{lin-2004-rouge}. Various ROUGE-N scores, depending on the number N of N-grams considered, are usually available in standard implementations \cite{Rouge-HF}. To ensure consistency, we used \textit{spaCy}'s rule-based sentencizer for sentence splitting. For ROUGE, we enabled the \textit{Porter Stemmer} \cite{Porter} internally to normalise words to their base forms (stems) while removing the additional morphemes (affixes).

A major limitation of the above class of metrics based on N-gram overlap is that they do not have any synonym or paraphrase awareness. The first metric incorporating synonym awareness was METEOR \cite{banerjee-lavie-2005-meteor} via the lexical resource WordNet Database \cite{10.7551/mitpress/7287.001.0001}. METEOR was introduced for machine translations, combining precision and recall in a harmonic mean, with more weight on the latter. A more modern approach to dealing with synonyms is to employ contextualised word embeddings, for example, the ones from BERT \cite{devlin-etal-2019-bert}.  The added value of these embeddings lies in the \textit{distributional hypothesis} approach \cite{Harris01081954}: words learnt in a similar context have similar meanings. BERTScore \cite{zhang2020bertscoreevaluatingtextgeneration} creates a metric out of BERT embeddings using pairwise cosine similarity and greedy matching. We adapted the standard BERTScore implementation to work with the ModernBERT architecture \cite{ModernBERT} and, as a checkpoint, we used the recently introduced \textit{BioClinical-ModernBERT-base} \cite{sounack2025bioclinicalmodernbertstateoftheartlongcontext}, which was further pretrained on a 53 million tokens biomedical and clinical corpus. We reported only the BERTScore-recall. We defer the study of the evolution of other metrics, such as COMET \cite{rei-etal-2020-comet} and BLEURT \cite{sellam-etal-2020-bleurt}, to future studies.

\section{Results}
\subsection{PEGASUS}
\begin{figure}[!ht]
  \centering
  \begin{minipage}[b]{0.48\textwidth}
    \centering
    \includegraphics[width=\textwidth]{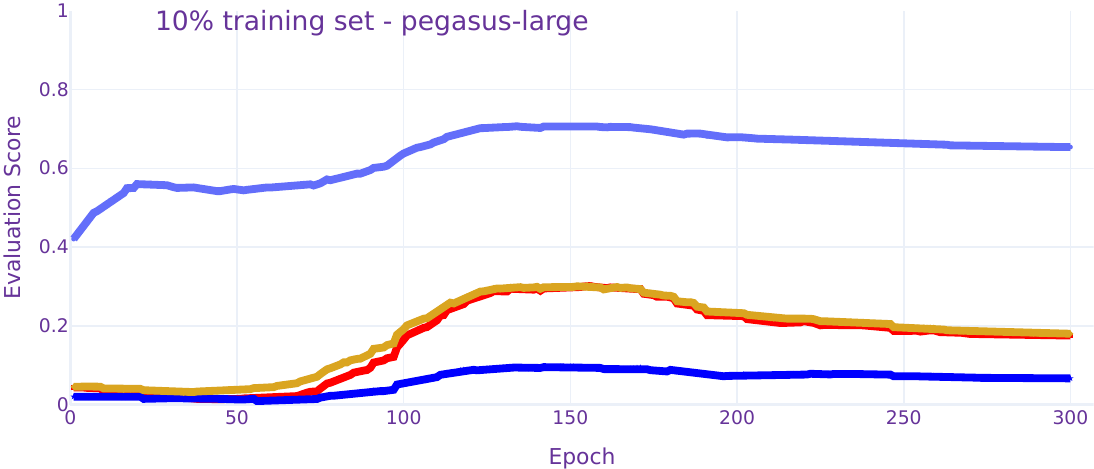}
  \end{minipage}
  \hspace{0.02\textwidth} 
  \begin{minipage}[b]{0.48\textwidth}
    \centering
    \includegraphics[width=\textwidth]{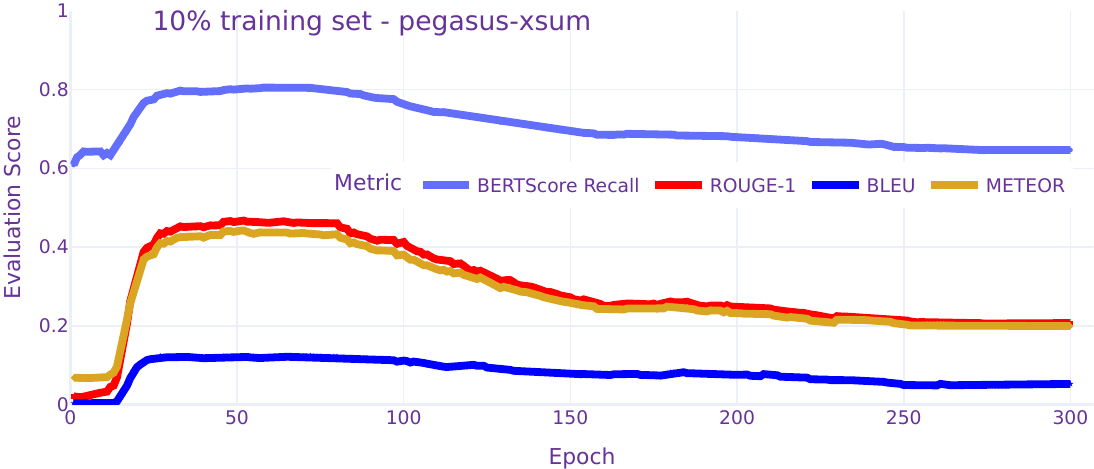}
  \end{minipage}

  \vspace{1em} 

  \begin{minipage}[b]{0.48\textwidth}
    \centering
    \includegraphics[width=\textwidth]{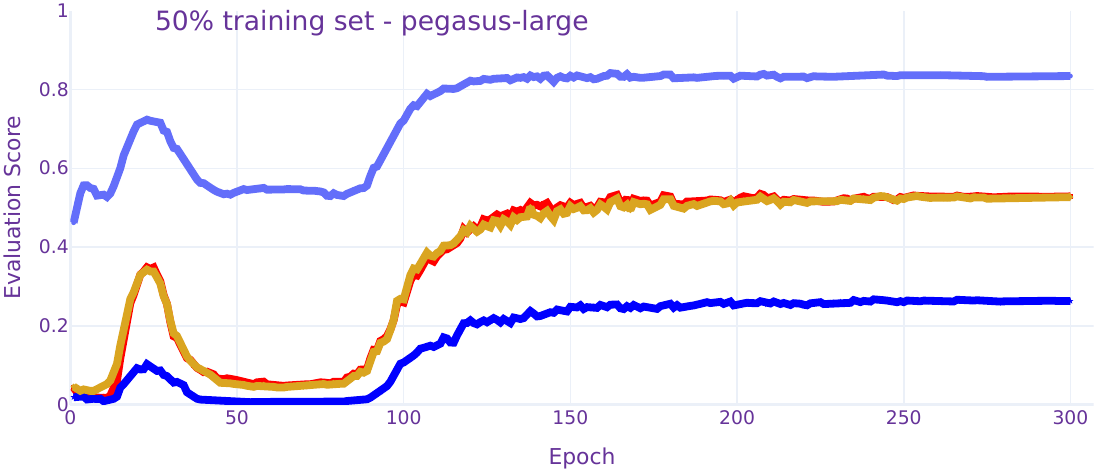}
  \end{minipage}
  \hspace{0.02\textwidth} 
  \begin{minipage}[b]{0.48\textwidth}
    \centering
    \includegraphics[width=\textwidth]{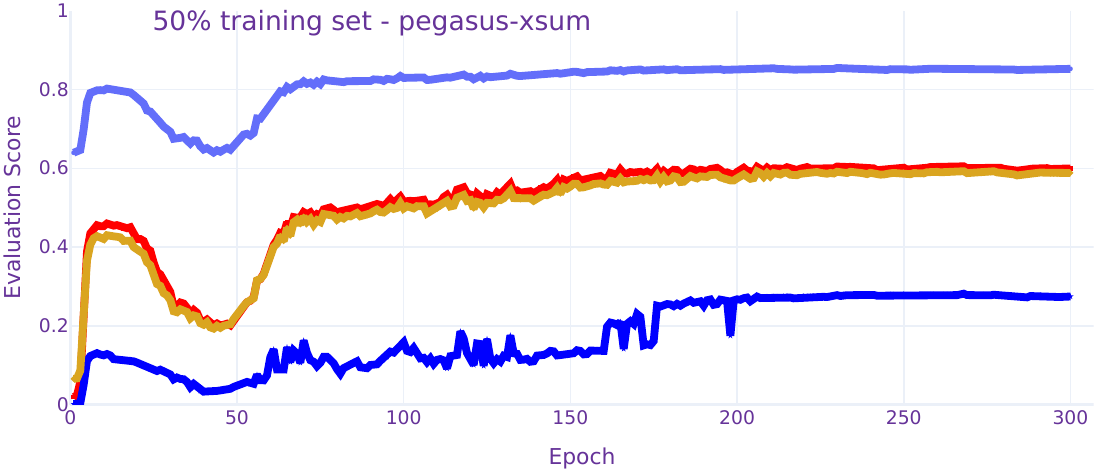}
  \end{minipage}

  \vspace{1em} 

  \begin{minipage}[b]{0.48\textwidth}
    \centering
    \includegraphics[width=\textwidth]{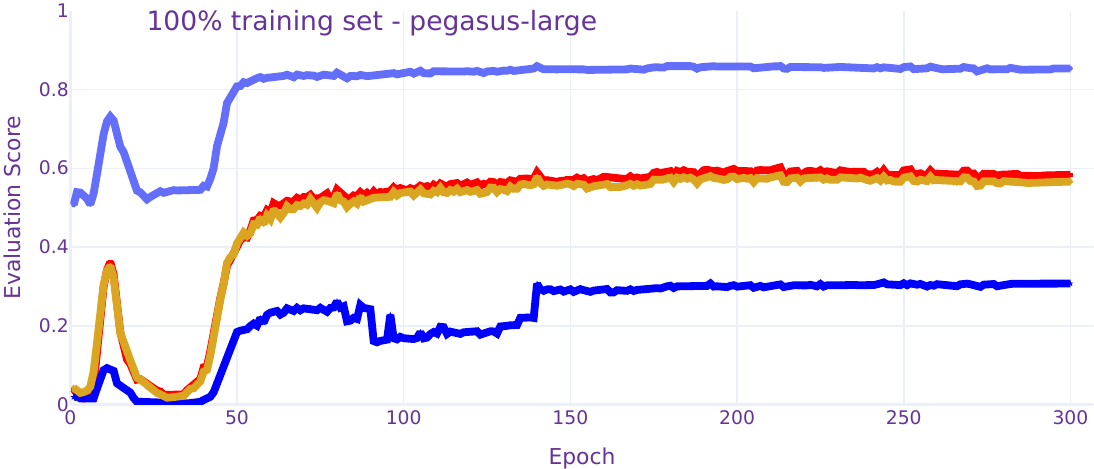}
  \end{minipage}
  \hspace{0.02\textwidth} 
  \begin{minipage}[b]{0.48\textwidth}
    \centering
    \includegraphics[width=\textwidth]{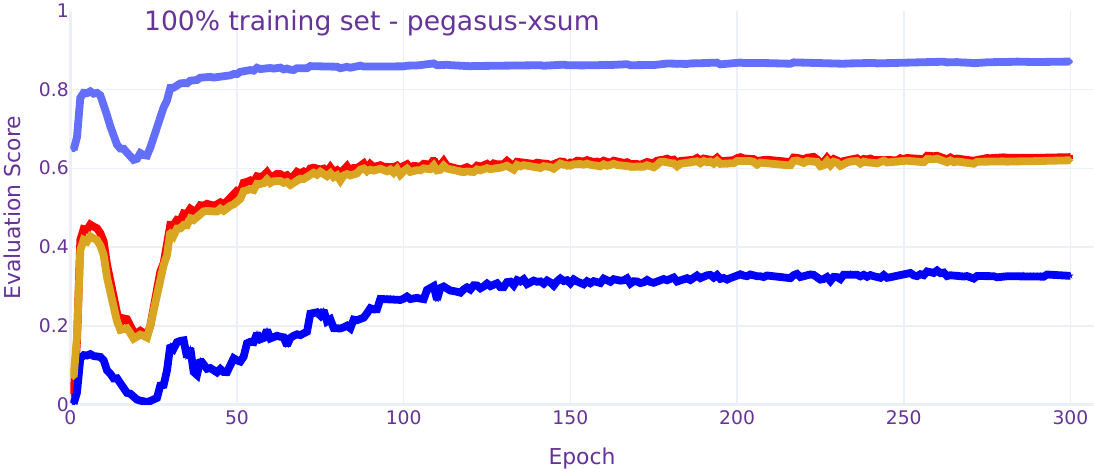}
  \end{minipage}
    \caption{Evolution of metrics scores on the validation set over epochs. Two flavours of PEGASUS models were used: \textit{large} (left) and \textit{xsum} (right). Training set sizes are in descending order. Metrics display an early peak, an intermediate phase where performance drops, and a saturation phase. This behaviour might indicate a loss of generalisation ability due to overfitting, a phenomenon known as \textit{catastrophic forgetting}. BLEU is a metric designed for machine translation: it is not surprising that it is significantly noisier and lower than the other measures, indicating its inability to evaluate summary qualities.} 
    \label{fig:eval_curves}
\end{figure}

The evolution of the metrics scores during the training of the PEGASUS model is shown in Fig. \ref{fig:eval_curves}.
In our experiments, we noticed a clear qualitative difference between the smallest fine-tuning training size (10\% of the original training split, i.e. 268 data points) and the other two (50\%, 100\%). 
In the 10\% simulations, both checkpoints exhibited a sharp rise in the metrics (peaking for \textit{large} at 155 epochs, \textit{xsum} at 52), followed by a significant degradation of them. This behaviour is not surprising: the model's expressivity is vastly superior to the variance of the (reduced) training set, eventually leading to \textit{overfitting}. Instead of acquiring generalisation ability, the model only learns to reproduce the training data and performs unsatisfyingly on the unseen data \cite{Goodfellow-et-al-2016}, in this case, the validation set. 

More interestingly, the metric scores relative to 50\% and 100\% of the training set sizes exhibited three distinct phases. 
At first, both checkpoints manifested an early peak (\textit{large} at 12 epochs, \textit{xsum} at 6 epochs), followed by a phase in which the scores dramatically dropped, similarly to what happened in the 10\% runs. We observed that the magnitude of the early peaks (identified as the local maximum of the ROUGE-1 score) is similar between all three training sizes, suggesting that the model's expressivity saturated the variance of the training dataset regardless of its size:  
\begin{table}[htbp!]
    \centering
    \begin{tabular}{ccccccc}
         Checkpoint & Train size & Peak (epochs) & ROUGE-1 & BERTScore-recall&  METEOR & BLEU \\
        \hline \hline
         \textit{large} & 10\% & 155 & \textbf{0.30}29 & 0.7076 & 0.3035 & 0.0950\\
         \textit{large}& 50\% & 23 & \textbf{0.34}89 & 0.7238 & 0.3429  & 0.1029\\
         \textit{large}& 100\% & 12 & \textbf{0.36}25 & 0.7315 & 0.3526 & 0.0900\\
         \hline
         \textit{xsum}& 10\% & 52& \textbf{0.46}76 & 0.8032 &  0.4423 & 0.1223\\
         \textit{xsum}& 50\% & 11 & \textbf{0.46}01 & 0.8027 & 0.4304  & 0.1286\\
         \textit{xsum}&  100\%& 6 & \textbf{0.45}74 & 0.7957 &  0.4275 & 0.1284\\
    \end{tabular}
    \caption{Metrics scores at the early-peak moment of the training history.}
    \label{tab:early_peaks}
\end{table}

Direct inspection of the outputs allowed us to qualitatively understand the model's behaviours during the first two phases: during the early-peaks, the models predicted either negated diagnoses (e.g., \textit{"No acute cardiopulmonary abnormality."}) or empty strings. Then, during the forgetting-phase, the models produced either negated diagnoses or jibberish (e.g: \textit{"Arrested Arrested..."}, \textit{"Lake Lake..."}, \textit{"Nurse Nurse..."}). Unlike the smallest training size, both checkpoints manifested a third, new, stationary phase in the later training history, wherein the metric scores surpassed those during the early peak. This onset happened for \textit{large}-50\% at 106 epochs, \textit{large}-100\% at 48, \textit{xsum}-50\% at 65, \textit{xsum}-100\% at 32.  In this phase, the models have finally acquired generalising abilities and were able to also generate  diagnoses (e.g. \textit{"Small left pleural effusion."}) on top of negated ones. We conjecture these three phases to be a manifestation of the more general phenomenon of the \textit{epoch-wise double-descent} behaviour in deep neural networks postulated in \cite{doi:10.1073/pnas.1903070116}, \cite{Nakkiran_2021}. 

In Table \ref{tab:early_peaks}, we noticed that the position of the early peaks w.r.t. epochs displayed inverse scaling with the train dataset size, leading us to assume that a generalisation ability could have been acquired for the smallest training sets, had they been continued for more than 530 (\textit{large}) and 330 epochs (\textit{xsum}). 
Additionally, we believe that \textit{xsum} performed better than \textit{large} because the first checkpoint is better apt for the concise summaries \cite{DemnerFushman2012DesignAD} typically used in chest X-ray reports.

As a complementary analysis, we rephrased the results of the checkpoints trained on 100\% of the training data as a binary classification task of negated diagnoses and diagnoses introduced in \ref{reference_cohort}. Results are shown in Table \ref{tab:binary_three_phases} and should be benchmarked against an always-negated diagnoses (dummy) classifier task, which, given the distribution of the two classes in the validation split, would have scored 0.62 precision and 1.0 recall. 
\begin{table}[htbp!]
    \centering
    \begin{tabular}{l||ccc|ccc}
    \multicolumn{1}{c||}{} &
    \multicolumn{3}{c|}{\textit{large} 100 \%} &
    \multicolumn{3}{c}{\textit{xsum} 100 \%} \\
    &  Epoch & Precision & Recall &   Epoch & Precision & Recall\\
    \hline
    \textbf{early peak }&  12 & 0.5579 & 0.6161 &  6 & \textbf{0.6242} & \textbf{0.9384}\\
    forgetting phase & 30 & 0.3077 & 0.019 &  23& 0.5965 & 0.3223\\
    generalisation& 200 & 0.8053 & 0.8626 & 200 & 0.7511 & 0.8009\\
    \end{tabular}
    \caption{Results of the PEGASUS checkpoints trained on the entire split as a binary classification task. These results should be read by keeping in mind that a dummy classifier would produce 0.62 precision and 1.0 of recall given the diagnoses distribution in the validation split.}
    \label{tab:binary_three_phases}
\end{table}

In particular, the predictions' deceptiveness during the early peak phase is well illustrated by the checkpoint \textit{xsum} achieving 0.62 as precision and 0.94 as recall. 

\subsection{PEGASUS-X}
\begin{figure}[htbp!]
  \centering
  \begin{minipage}[b]{0.48\textwidth}
    \centering
    \includegraphics[width=\textwidth]{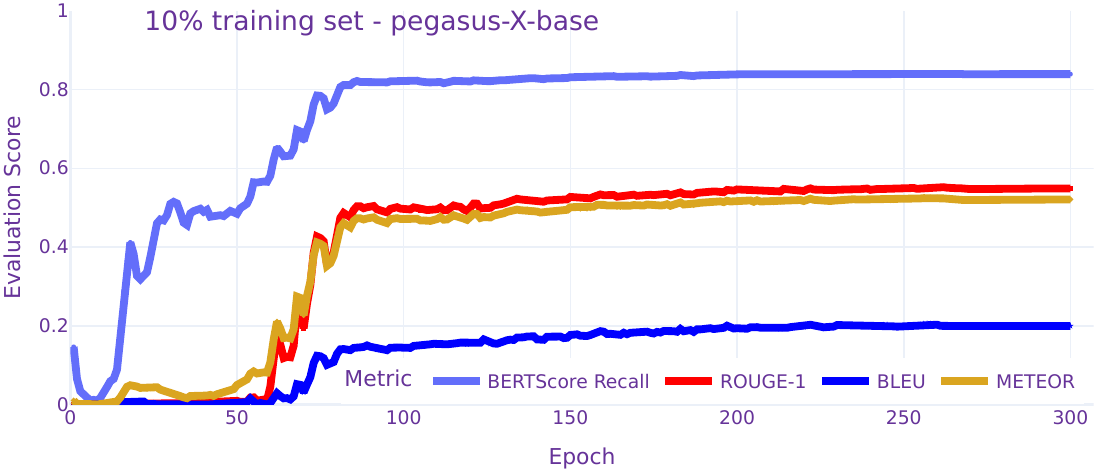}
  \end{minipage}
  \hspace{0.02\textwidth} 
  \begin{minipage}[b]{0.48\textwidth}
    \centering
    \includegraphics[width=\textwidth]{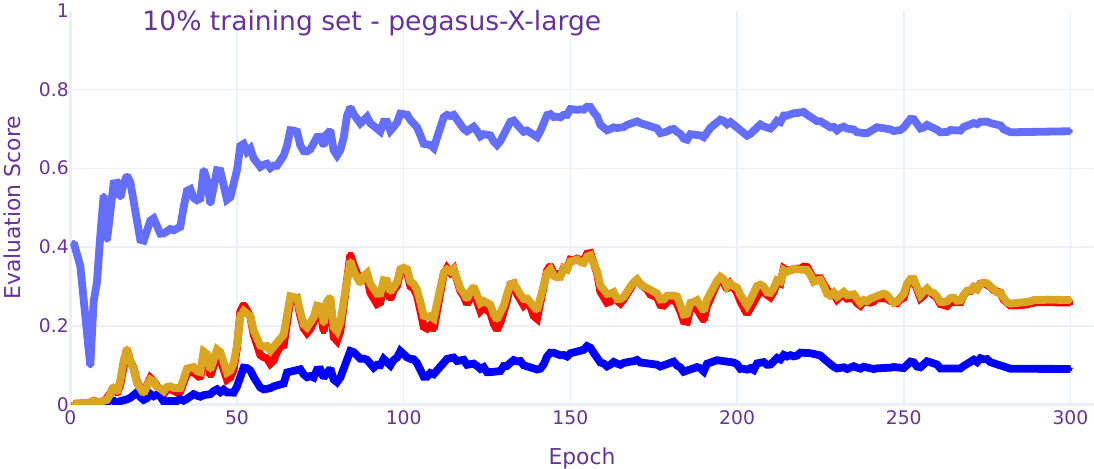}
  \end{minipage}

  \vspace{1em} 

  \begin{minipage}[b]{0.48\textwidth}
    \centering
    \includegraphics[width=\textwidth]{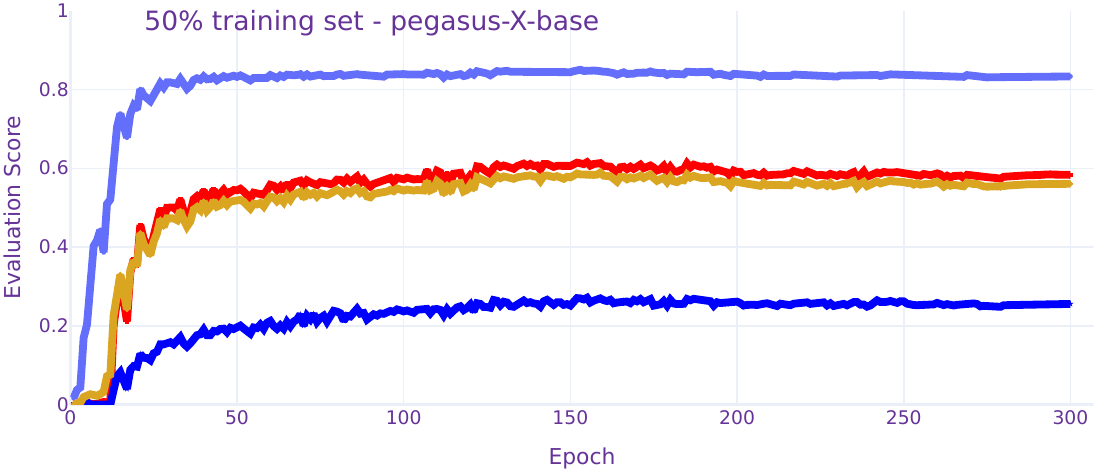}
  \end{minipage}
  \hspace{0.02\textwidth} 
  \begin{minipage}[b]{0.48\textwidth}
    \centering
    \includegraphics[width=\textwidth]{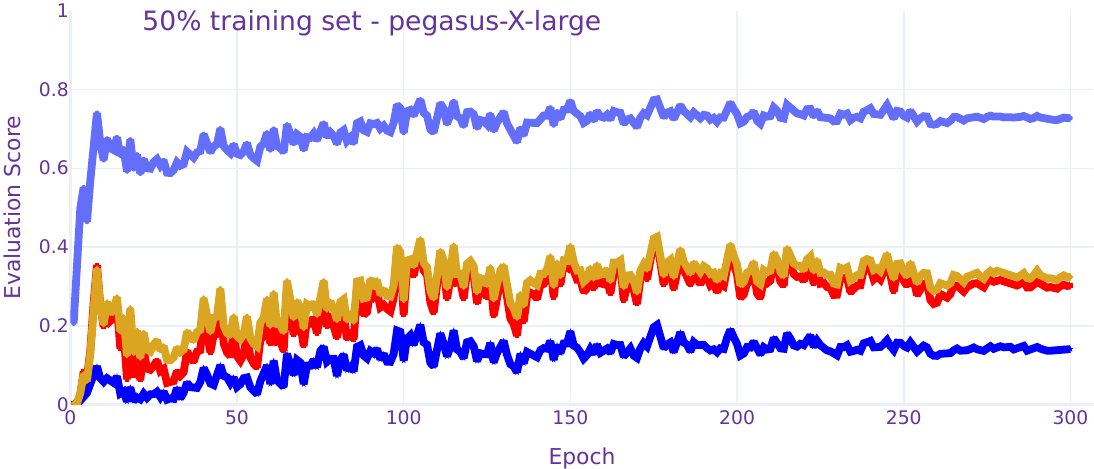}
  \end{minipage}

  \vspace{1em} 

  \begin{minipage}[b]{0.48\textwidth}
    \centering
    \includegraphics[width=\textwidth]{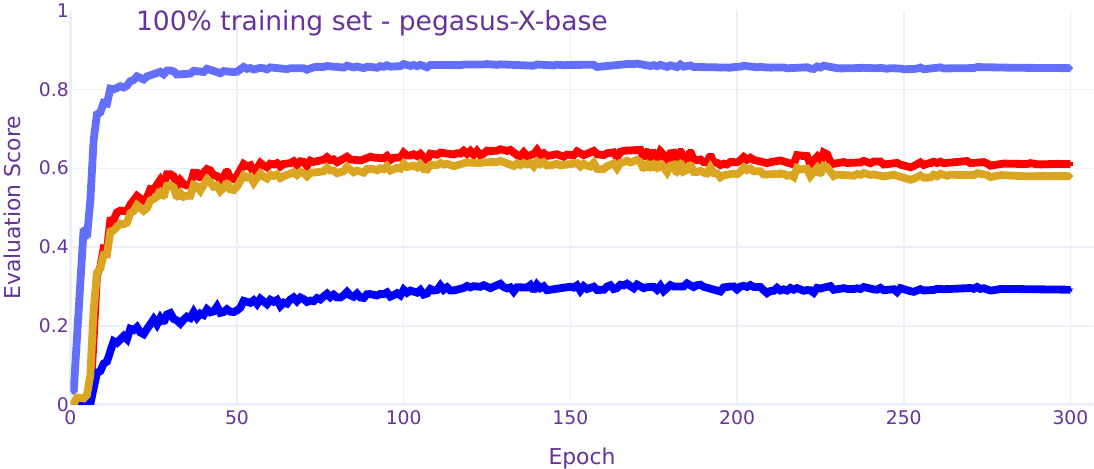}
  \end{minipage}
  \hspace{0.02\textwidth} 
  \begin{minipage}[b]{0.48\textwidth}
    \centering
    \includegraphics[width=\textwidth]{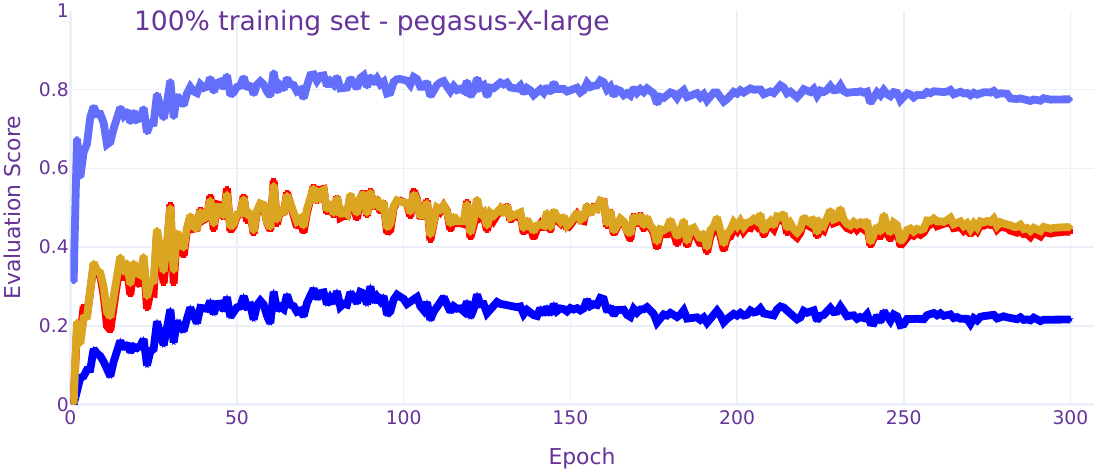}
  \end{minipage}
    \caption{Evolution of metrics scores on the validation set over epochs. Two sizes of PEGASUS-X models were used: \textit{base} (left) and \textit{large} (right). Training set sizes are in descending order.} 
    \label{fig:eval_curves_PegasusX}

\end{figure}

We used for PEGASUS-X two checkpoints that differ by the number of parameters and not by the pretraining set: \textit{base} has 257 million parameters, whereas \textit{large} has 512 million. Their behaviour is significantly different from what we observed for PEGASUS, see Fig. \ref{fig:eval_curves_PegasusX}. 

The smaller checkpoint displayed a nearly-monotonous increase in the metrics over epochs, with the performances plateauing  faster with larger training data size, similarly to what we observed in PEGASUS. The 95\%\footnote{This threshold is chosen arbitrarily to avoid taking care of the noise in the plateau phase} of the ROUGE-1 score at 300 epochs is reached in fact for 10\% of the data at 134 epochs, for 50\% at 60 epochs, for 100\% at 29 epochs.

The \textit{large} checkpoint exhibited a more puzzling behaviour. The metric scores were lower than the ones for the smaller checkpoint, and the training history was jagged, which is a typical symptom of overfitting \cite{Nakkiran_2021}. 
In this case, the model was unable to understand the task, and the outputs were combinations of the pre-training datasets and the fine-tuning data (e.g.: \textit{"Midfielder of Hungary's Ferencvarous"}, \textit{"goals scored with right upper lobe pneumonia."}, \textit{"Everton defenders are grossly clear of left lung effusion."}). These observations suggest that the usage of a model with different pretraining objectives (very short summaries versus long summaries) and too many parameters with respect to the amount of fine-tuning data hurts the performance and might not be mitigated even with more epochs.

\section{Discussion}
In this work, we showed the challenges that fine-tuning of large, expressive models poses in the typical clinical setting of scarcity of labelled training data. Rather than decoder models (LLMs), we picked two encoder-decoder architectures specialised for summarisation, PEGASUS and PEGASUS-X. For each model's family, we fine-tuned and evaluated two checkpoints on the training and validation split of a public medium-sized radiological reports dataset. The PEGASUS experiments showed that naive fine-tuning on such a limited-variability dataset displays catastrophic forgetting and peak-and-drop recovery. Such a phenomenon is not peculiar to the encoder-decoder model family \cite{raffel2020exploring} \cite{lewis2019bart}, but it has been observed in all three transformer families \cite{chang2025lora} \cite{mosbach2020stability}. Our advice to clinicians is to carefully monitor the training history (evolution of metrics over epochs), because aggressive early-stopping techniques might actually hinder the search for the best model during fine-tuning due to the possible existence of an early peak.
Complementary techniques to curb this pitfall could be showing the models examples of increasing difficulty (curriculum learning \cite{gururangan2020dontstoppretrainingadapt} \cite{chen2020recalllearnfinetuningdeep}), elastic weight consolidation, layer unfreezing, regularisation techniques, or employing overfitting alarms transformer models that treat the training history as a time series problem \cite{Li_2024}. Another approach could be to rely on Parameter Efficient Fine-Tuning (PEFT) techniques \cite{hu2021loralowrankadaptationlarge} \cite{dettmers2023qloraefficientfinetuningquantized} at the price of adding additional hyperparameters, and the related hyperparameter optimisation. 

The experiments regarding PEGASUS-X showed that using a larger checkpoint does not always improve the performance. In this case, the model did not only face the challenge of limited fine-tuning data, but also of having a different pretraining objective. In fact, PEGASUS-X is a PEGASUS model adapted for long texts (up to 16384 tokens instead of 512), while the data it was tested on was composed of very short sentences, see Fig. \ref{fig:preprocessed_dataset}.

As it is common for summarisation tasks, we hereby report the best ROUGE-1 scores for all the models/checkpoints, and the corresponding epoch. All of the best results were obtained with 100\% of the training data. 

\begin{table}[htbp!]
\centering 
\begin{tabular}{c|c|c|c|c|}
model/checkpoint & \textit{pegasus/large} & \textit{pegasus/xsum} & \textit{pegasus-x/base} & \textit{pegasus-x/large}  \\ 
\hline
ROUGE-1 & 0.6027 & 0.6323 & \textbf{0.6505} & 0.5687  \\
\hline
epoch & 213 & 257 & 170 & 61
\end{tabular}
\end{table}
We defer to future work to compare, with the help of trained clinicians, the most representative fine-tuned models on the test split.

\section{Conclusion}
In this work, we aimed to provide insights on how to adapt non-domain-specific abstractive summarisation models and, at the same time, to give actionable insights to practitioners on how to avoid over- and underfitting. Our findings confirmed that naive fine-tuning on a data set of limited variability can cause catastrophic forgetting. Consequently, we advise against techniques of aggressive early-stopping. Also, we want to point out that, in such a scenario, the adaptation of larger models can result in lower performance. In both cases, we want to highlight the critical need for clinicians to carefully monitor the full training history to find the optimally trained checkpoint. We acknowledge the limitation that the dataset used in this study is of a limited size and that our investigation focused solely on one specific model family, which affects the generalisability of our results. 

Furthermore, automatic evaluation in such \textit{seq2seq} tasks as summarisation is known to be problematic \cite{10.1162/tacl_a_00373} \cite{Chauhan2022ACS}, \cite{10.1145/3485766}, and correlation with human judgement remains the most reliable way of assessing the output quality \cite{krubinski-etal-2021-just}. Future research is needed to validate our findings in a different/wider range of datasets and with other and more recent model architectures. 
Our work lays the groundwork for future investigations into more robust fine-tuning strategies for summarisation models in specialised domains.

\section{Data Availability}
The dataset is publicly accessible at \cite{van_veen_adapted_2024_github}.
\section{Code availability}
The code can be made available by reasonable request.

\section{Author contributions}
C.B. conceived the study. C.B. conducted data extraction. C.B. performed statistical and machine learning modelling. C.B. and M.L. analysed the data. C.B. created the figures and tables. C.B., M.L., M.E.M., and M.B. wrote the manuscript. M.E.M. advised clinical aspects of the study. M.B. and L.M. acquired funding and supervised the study. All authors critically reviewed and approved the final manuscript version. 
\section{Acknowledgments}
C.B. and L.M. acknowledge funding from the
German Federal Ministry of Education and Research (BMBF) within the Junior Research Group "Klinische Textanalytik: Methoden f\"ur NLP an deutschen Texten" (DE.xt, FKZ: 01ZZ2009). M.L. and M.E.M. acknowledge funding from the (BMBF) within the framework of the MII: Medical Informatics in Research and Care in University Medicine (MIRACUM) Consortium for the Junior Research Group "Medical informatics for holistic disease models in personalised and preventive medicine" (MIDorAI, FKZ: 01ZZ2020). The funders had no role in study design, data collection, analysis, interpretation, the decision to publish, or preparation of the manuscript.
C.B. acknowledges fruitful discussions with S. Bolo. 
\section{Competing interests}
M.E.M. reports unrelated consultancy to EppData GmbH and Siemens Healthineers GmbH, Germany. 
The other authors declare no competing interests.
\clearpage
\bibliographystyle{vancouver}
\bibliography{literature}

\begin{thebibliography}{10}

\bibitem{van_veen_adapted_2024}
Van~Veen D, Van~Uden C, Blankemeier L, Delbrouck JB, Aali A, Bluethgen C,
  et~al.
\newblock Adapted large language models can outperform medical experts in
  clinical text summarization.
\newblock Nature Medicine. 2024 Apr;30(4):1134-42.
\newblock Available from:
  \url{https://www.nature.com/articles/s41591-024-02855-5}.

\bibitem{Zhang2019PEGASUSPW}
Zhang J, Zhao Y, Saleh M, Liu PJ.
\newblock PEGASUS: Pre-training with Extracted Gap-sentences for Abstractive
  Summarization.
\newblock ArXiv. 2019;abs/1912.08777.
\newblock Available from:
  \url{https://api.semanticscholar.org/CorpusID:209405420}.

\bibitem{phang2022investigating}
Phang J, Zhao Y, Liu PJ.
\newblock Investigating efficiently extending transformers for long input
  summarization.
\newblock arXiv preprint arXiv:220804347. 2022.

\bibitem{nishio2024fully}
Nishio M, Matsunaga T, Matsuo H, Nogami M, Kurata Y, Fujimoto K, et~al.
\newblock Fully automatic summarization of radiology reports using natural
  language processing with large language models.
\newblock Informatics in Medicine Unlocked. 2024;46:101465.

\bibitem{schutze2008introduction}
Sch{\"u}tze H, Manning CD, Raghavan P.
\newblock Introduction to information retrieval. vol.~39.
\newblock Cambridge University Press Cambridge; 2008.

\bibitem{pons2016natural}
Pons E, Braun LM, Hunink MM, Kors JA.
\newblock Natural language processing in radiology: a systematic review.
\newblock Radiology. 2016;279(2):329-43.

\bibitem{maros2021comparative}
Maros ME, Cho CG, Junge AG, K{\"a}mpgen B, Saase V, Siegel F, et~al.
\newblock Comparative analysis of machine learning algorithms for
  computer-assisted reporting based on fully automated cross-lingual RadLex
  mappings.
\newblock Scientific Reports. 2021;11(1):5529.

\bibitem{doi:10.1148/radiol.2017171115}
Chen MC, Ball RL, Yang L, Moradzadeh N, Chapman BE, Larson DB, et~al.
\newblock Deep Learning to Classify Radiology Free-Text Reports.
\newblock Radiology. 2018;286(3):845-52.
\newblock PMID: 29135365.
\newblock Available from: \url{https://doi.org/10.1148/radiol.2017171115}.

\bibitem{CHAPMAN2011728}
Chapman BE, Lee S, Kang HP, Chapman WW.
\newblock Document-level classification of CT pulmonary angiography reports
  based on an extension of the ConText algorithm.
\newblock Journal of Biomedical Informatics. 2011;44(5):728-37.
\newblock Available from:
  \url{https://www.sciencedirect.com/science/article/pii/S1532046411000621}.

\bibitem{vaswani2017attention}
Vaswani A, Shazeer N, Parmar N, Uszkoreit J, Jones L, Gomez AN, et~al.
\newblock Attention is all you need.
\newblock Advances in neural information processing systems. 2017;30.

\bibitem{2018arXiv181004805D}
{Devlin} J, {Chang} MW, {Lee} K, {Toutanova} K.
\newblock {BERT: Pre-training of Deep Bidirectional Transformers for Language
  Understanding}.
\newblock arXiv e-prints. 2018 Oct:arXiv:1810.04805.

\bibitem{warner2024smarterbetterfasterlonger}
Warner B, Chaffin A, Clavié B, Weller O, Hallström O, Taghadouini S, et~al..
  Smarter, Better, Faster, Longer: A Modern Bidirectional Encoder for Fast,
  Memory Efficient, and Long Context Finetuning and Inference; 2024.
\newblock Available from: \url{https://arxiv.org/abs/2412.13663}.

\bibitem{ModernBERT}
Warner B, Chaffin A, Clavié B, Weller O, Hallström O, Taghadouini S, et~al..
  Finally, a Replacement for BERT; 2024.
\newblock \url{https://huggingface.co/blog/modernbert}.

\bibitem{radford2019language}
Radford A, Wu J, Child R, Luan D, Amodei D, Sutskever I, et~al.
\newblock Language models are unsupervised multitask learners.
\newblock OpenAI blog. 2019;1(8):9.

\bibitem{2020arXiv200514165B}
{Brown} TB, {Mann} B, {Ryder} N, {Subbiah} M, {Kaplan} J, {Dhariwal} P, et~al.
\newblock {Language Models are Few-Shot Learners}.
\newblock arXiv e-prints. 2020 May:arXiv:2005.14165.

\bibitem{openai2024gpt4technicalreport}
OpenAI, Achiam J, Adler S, Agarwal S, Ahmad L, Akkaya I, et~al.. GPT-4
  Technical Report; 2024.
\newblock Available from: \url{https://arxiv.org/abs/2303.08774}.

\bibitem{decoder-only}
Bai Y. Why are most LLMs decoder-only?;.
\newblock Accessed: 2025-03-21.
\newblock
  \url{https://medium.com/@yumo-bai/why-are-most-llms-decoder-only-590c903e4789}.

\bibitem{10.1162/neco.1997.9.8.1735}
Hochreiter S, Schmidhuber J.
\newblock Long Short-Term Memory.
\newblock Neural Computation. 1997 11;9(8):1735-80.
\newblock Available from: \url{https://doi.org/10.1162/neco.1997.9.8.1735}.

\bibitem{chaves2023rales}
Chaves JMZ, Bhaskhar N, Attias M, Delbrouck JB, Rubin D, Loening AM, et~al.
\newblock Ra{LE}s: a Benchmark for Radiology Language Evaluations.
\newblock In: Thirty-seventh Conference on Neural Information Processing
  Systems Datasets and Benchmarks Track; 2023. Available from:
  \url{https://openreview.net/forum?id=PWLGrvoqiR}.

\bibitem{jurafsky2025speech}
Jurafsky D, Martin HJ.
\newblock Speech and Language Processing: An Introduction to Natural Language
  Processing, Computational Linguistics, and Speech Recognition with Language
  Models, 3rd edition; 2025.
\newblock Available from: \url{https://web.stanford.edu/~jurafsky/slp3}.

\bibitem{dai-etal-2021-bdkg}
Dai S, Wang Q, Lyu Y, Zhu Y.
\newblock {BDKG} at {MEDIQA} 2021: System Report for the Radiology Report
  Summarization Task.
\newblock In: Demner-Fushman D, Cohen KB, Ananiadou S, Tsujii J, editors.
  Proceedings of the 20th Workshop on Biomedical Language Processing. Online:
  Association for Computational Linguistics; 2021. p. 103-11.
\newblock Available from: \url{https://aclanthology.org/2021.bionlp-1.11/}.

\bibitem{MEDIQA}
MEDIQA 2021 Summarization in the Medical Domain;.
\newblock Available from: \url{https://github.com/abachaa/MEDIQA2021}.

\bibitem{jaeger2013automatic}
Jaeger S, Karargyris A, Candemir S, Folio L, Siegelman J, Callaghan F, et~al.
\newblock Automatic tuberculosis screening using chest radiographs.
\newblock IEEE transactions on medical imaging. 2013;33(2):233-45.

\bibitem{open-i}
Demner-Fushman D, Kohli MD, Rosenman MB, Shooshan SE, Rodriguez L, Antani S,
  et~al.
\newblock Preparing a collection of radiology examinations for distribution and
  retrieval.
\newblock Journal of the American Medical Informatics Association. 2015
  07;23(2):304-10.
\newblock Available from: \url{https://doi.org/10.1093/jamia/ocv080}.

\bibitem{van_veen_adapted_2024_github}
Van~Veen D, Van~Uden C, Blankemeier L, Delbrouck JB, Aali A, Bluethgen C,
  et~al.. Adapted large language models can outperform medical experts in
  clinical text summarization. GitHub; 2024.
\newblock Accessed: 2025-06-03.
\newblock \url{https://github.com/StanfordMIMI/clin-summ}.

\bibitem{kocak2023checklist}
Kocak B, Baessler B, Bakas S, Cuocolo R, Fedorov A, Maier-Hein L, et~al.
\newblock CheckList for EvaluAtion of Radiomics research (CLEAR): a
  step-by-step reporting guideline for authors and reviewers endorsed by ESR
  and EuSoMII.
\newblock Insights into imaging. 2023;14(1):75.

\bibitem{bird2009natural}
Bird S, Klein E, Loper E.
\newblock Natural language processing with Python: analyzing text with the
  natural language toolkit.
\newblock " O'Reilly Media, Inc."; 2009.

\bibitem{huggingfaceConcept}
{HuggingFace}. Padding and truncation. HuggingFace;.
\newblock Accessed: 2025-03-06.
\newblock \url{https://huggingface.co/docs/transformers/pad_truncation}.

\bibitem{MLPlusMahalanobis}
{Machine Learning Plus}. Mahalanobis Distance: Definition, Formula, and
  Examples;.
\newblock Accessed: 2025-03-10.
\newblock
  \url{https://www.machinelearningplus.com/statistics/mahalanobis-distance/}.

\bibitem{10.1002/widm.1236}
Rousseeuw PJ, Hubert M.
\newblock Anomaly detection by robust statistics.
\newblock WIREs Data Mining and Knowledge Discovery. 2017;8.

\bibitem{narayan-etal-2018-dont}
Narayan S, Cohen SB, Lapata M.
\newblock Don`t Give Me the Details, Just the Summary! Topic-Aware
  Convolutional Neural Networks for Extreme Summarization.
\newblock In: Riloff E, Chiang D, Hockenmaier J, Tsujii J, editors. Proceedings
  of the 2018 Conference on Empirical Methods in Natural Language Processing.
  Brussels, Belgium: Association for Computational Linguistics; 2018. p.
  1797-807.
\newblock Available from: \url{https://aclanthology.org/D18-1206/}.

\bibitem{gururangan2020dontstoppretrainingadapt}
Gururangan S, Marasović A, Swayamdipta S, Lo K, Beltagy I, Downey D, et~al..
  Don't Stop Pretraining: Adapt Language Models to Domains and Tasks; 2020.
\newblock Available from: \url{https://arxiv.org/abs/2004.10964}.

\bibitem{kudo-richardson-2018-sentencepiece}
Kudo T, Richardson J.
\newblock {S}entence{P}iece: A simple and language independent subword
  tokenizer and detokenizer for Neural Text Processing.
\newblock In: Blanco E, Lu W, editors. Proceedings of the 2018 Conference on
  Empirical Methods in Natural Language Processing: System Demonstrations.
  Brussels, Belgium: Association for Computational Linguistics; 2018. p. 66-71.
\newblock Available from: \url{https://aclanthology.org/D18-2012/}.

\bibitem{sennrich-etal-2016-neural}
Sennrich R, Haddow B, Birch A.
\newblock Neural Machine Translation of Rare Words with Subword Units.
\newblock In: Erk K, Smith NA, editors. Proceedings of the 54th Annual Meeting
  of the Association for Computational Linguistics (Volume 1: Long Papers).
  Berlin, Germany: Association for Computational Linguistics; 2016. p. 1715-25.
\newblock Available from: \url{https://aclanthology.org/P16-1162/}.

\bibitem{DBLP:journals/corr/abs-1804-10959}
Kudo T.
\newblock Subword Regularization: Improving Neural Network Translation Models
  with Multiple Subword Candidates.
\newblock CoRR. 2018;abs/1804.10959.
\newblock Available from: \url{http://arxiv.org/abs/1804.10959}.

\bibitem{075rule-of-thumb}
{OpenAI}. What are tokens and how to count them?;.
\newblock Accessed: 2025-03-10.
\newblock
  \url{https://help.openai.com/en/articles/4936856-what-are-tokens-and-how-to-count-them}.

\bibitem{10.1145/3485766}
Sai AB, Mohankumar AK, Khapra MM.
\newblock A Survey of Evaluation Metrics Used for NLG Systems.
\newblock ACM Comput Surv. 2022 Jan;55(2).
\newblock Available from: \url{https://doi.org/10.1145/3485766}.

\bibitem{papineni-etal-2002-bleu}
Papineni K, Roukos S, Ward T, Zhu WJ.
\newblock {B}leu: a Method for Automatic Evaluation of Machine Translation.
\newblock In: Isabelle P, Charniak E, Lin D, editors. Proceedings of the 40th
  Annual Meeting of the Association for Computational Linguistics.
  Philadelphia, Pennsylvania, USA: Association for Computational Linguistics;
  2002. p. 311-8.
\newblock Available from: \url{https://aclanthology.org/P02-1040/}.

\bibitem{post-2018-call}
Post M.
\newblock A Call for Clarity in Reporting {BLEU} Scores.
\newblock In: Proceedings of the Third Conference on Machine Translation:
  Research Papers. Belgium, Brussels: Association for Computational
  Linguistics; 2018. p. 186-91.
\newblock Available from: \url{https://www.aclweb.org/anthology/W18-6319}.

\bibitem{lin-2004-rouge}
Lin CY.
\newblock {ROUGE}: A Package for Automatic Evaluation of Summaries.
\newblock In: Text Summarization Branches Out. Barcelona, Spain: Association
  for Computational Linguistics; 2004. p. 74-81.
\newblock Available from: \url{https://www.aclweb.org/anthology/W04-1013}.

\bibitem{Rouge-HF}
{HuggingFace}. Metric: rouge. HuggingFace;.
\newblock Accessed: 2025-03-24.
\newblock \url{https://huggingface.co/spaces/evaluate-metric/rouge}.

\bibitem{Porter}
Porter MF.
\newblock An algoritm for suffix stripping.
\newblock Program. 2006 07;14:130-7.

\bibitem{banerjee-lavie-2005-meteor}
Banerjee S, Lavie A.
\newblock {METEOR}: An Automatic Metric for {MT} Evaluation with Improved
  Correlation with Human Judgments.
\newblock In: Goldstein J, Lavie A, Lin CY, Voss C, editors. Proceedings of the
  {ACL} Workshop on Intrinsic and Extrinsic Evaluation Measures for Machine
  Translation and/or Summarization. Ann Arbor, Michigan: Association for
  Computational Linguistics; 2005. p. 65-72.
\newblock Available from: \url{https://aclanthology.org/W05-0909/}.

\bibitem{10.7551/mitpress/7287.001.0001}
Fellbaum C.
\newblock WordNet: An Electronic Lexical Database.
\newblock The MIT Press; 1998.
\newblock Available from: \url{https://doi.org/10.7551/mitpress/7287.001.0001}.

\bibitem{devlin-etal-2019-bert}
Devlin J, Chang MW, Lee K, Toutanova K.
\newblock {BERT}: Pre-training of Deep Bidirectional Transformers for Language
  Understanding.
\newblock In: Burstein J, Doran C, Solorio T, editors. Proceedings of the 2019
  Conference of the North {A}merican Chapter of the Association for
  Computational Linguistics: Human Language Technologies, Volume 1 (Long and
  Short Papers). Minneapolis, Minnesota: Association for Computational
  Linguistics; 2019. p. 4171-86.
\newblock Available from: \url{https://aclanthology.org/N19-1423/}.

\bibitem{Harris01081954}
and ZSH.
\newblock Distributional Structure.
\newblock WORD. 1954;10(2-3):146-62.
\newblock Available from: \url{https://doi.org/10.1080/00437956.1954.11659520}.

\bibitem{zhang2020bertscoreevaluatingtextgeneration}
Zhang T, Kishore V, Wu F, Weinberger KQ, Artzi Y. BERTScore: Evaluating Text
  Generation with BERT; 2020.
\newblock Available from: \url{https://arxiv.org/abs/1904.09675}.

\bibitem{sounack2025bioclinicalmodernbertstateoftheartlongcontext}
Sounack T, Davis J, Durieux B, Chaffin A, Pollard TJ, Lehman E, et~al..
  BioClinical ModernBERT: A State-of-the-Art Long-Context Encoder for
  Biomedical and Clinical NLP; 2025.
\newblock Available from: \url{https://arxiv.org/abs/2506.10896}.

\bibitem{rei-etal-2020-comet}
Rei R, Stewart C, Farinha AC, Lavie A.
\newblock {COMET}: A Neural Framework for {MT} Evaluation.
\newblock In: Webber B, Cohn T, He Y, Liu Y, editors. Proceedings of the 2020
  Conference on Empirical Methods in Natural Language Processing (EMNLP).
  Online: Association for Computational Linguistics; 2020. p. 2685-702.
\newblock Available from: \url{https://aclanthology.org/2020.emnlp-main.213/}.

\bibitem{sellam-etal-2020-bleurt}
Sellam T, Das D, Parikh A.
\newblock {BLEURT}: Learning Robust Metrics for Text Generation.
\newblock In: Jurafsky D, Chai J, Schluter N, Tetreault J, editors. Proceedings
  of the 58th Annual Meeting of the Association for Computational Linguistics.
  Online: Association for Computational Linguistics; 2020. p. 7881-92.
\newblock Available from: \url{https://aclanthology.org/2020.acl-main.704/}.

\bibitem{Goodfellow-et-al-2016}
Goodfellow I, Bengio Y, Courville A.
\newblock Deep Learning.
\newblock MIT Press; 2016.
\newblock \url{http://www.deeplearningbook.org}.

\bibitem{doi:10.1073/pnas.1903070116}
Belkin M, Hsu D, Ma S, Mandal S.
\newblock Reconciling modern machine-learning practice and the classical
  bias–variance trade-off.
\newblock Proceedings of the National Academy of Sciences.
  2019;116(32):15849-54.
\newblock Available from:
  \url{https://www.pnas.org/doi/abs/10.1073/pnas.1903070116}.

\bibitem{Nakkiran_2021}
Nakkiran P, Kaplun G, Bansal Y, Yang T, Barak B, Sutskever I.
\newblock Deep double descent: where bigger models and more data hurt*.
\newblock Journal of Statistical Mechanics: Theory and Experiment. 2021
  dec;2021(12):124003.
\newblock Available from: \url{https://dx.doi.org/10.1088/1742-5468/ac3a74}.

\bibitem{DemnerFushman2012DesignAD}
Demner-Fushman D, Antani SK, Simpson MS, Thoma GR.
\newblock Design and Development of a Multimodal Biomedical Information
  Retrieval System.
\newblock J Comput Sci Eng. 2012;6:168-77.
\newblock Available from:
  \url{https://api.semanticscholar.org/CorpusID:6179464}.

\bibitem{raffel2020exploring}
Raffel C, Shazeer N, Roberts A, Lee K, Narang S, Matena M, et~al.
\newblock Exploring the limits of transfer learning with a unified text-to-text
  transformer.
\newblock Journal of machine learning research. 2020;21(140):1-67.

\bibitem{lewis2019bart}
Lewis M, Liu Y, Goyal N, Ghazvininejad M, Mohamed A, Levy O, et~al.
\newblock BART: Denoising sequence-to-sequence pre-training for natural
  language generation, translation, and comprehension.
\newblock arXiv preprint arXiv:191013461. 2019.

\bibitem{chang2025lora}
Chang Y, Guo C, Chang Y, Wu Y.
\newblock LoRA-GGPO: Mitigating Double Descent in LoRA Fine-Tuning via
  Gradient-Guided Perturbation Optimization.
\newblock arXiv preprint arXiv:250214538. 2025.

\bibitem{mosbach2020stability}
Mosbach M, Andriushchenko M, Klakow D.
\newblock On the stability of fine-tuning bert: Misconceptions, explanations,
  and strong baselines.
\newblock arXiv preprint arXiv:200604884. 2020.

\bibitem{chen2020recalllearnfinetuningdeep}
Chen S, Hou Y, Cui Y, Che W, Liu T, Yu X. Recall and Learn: Fine-tuning Deep
  Pretrained Language Models with Less Forgetting; 2020.
\newblock Available from: \url{https://arxiv.org/abs/2004.12651}.

\bibitem{Li_2024}
Li H, Rajbahadur GK, Lin D, Bezemer CP, Jiang ZM.
\newblock Keeping Deep Learning Models in Check: A History-Based Approach to
  Mitigate Overfitting.
\newblock IEEE Access. 2024;12:70676–70689.
\newblock Available from: \url{http://dx.doi.org/10.1109/ACCESS.2024.3402543}.

\bibitem{hu2021loralowrankadaptationlarge}
Hu EJ, Shen Y, Wallis P, Allen-Zhu Z, Li Y, Wang S, et~al.. LoRA: Low-Rank
  Adaptation of Large Language Models; 2021.
\newblock Available from: \url{https://arxiv.org/abs/2106.09685}.

\bibitem{dettmers2023qloraefficientfinetuningquantized}
Dettmers T, Pagnoni A, Holtzman A, Zettlemoyer L. QLoRA: Efficient Finetuning
  of Quantized LLMs; 2023.
\newblock Available from: \url{https://arxiv.org/abs/2305.14314}.

\bibitem{10.1162/tacl_a_00373}
Fabbri AR, Kryściński W, McCann B, Xiong C, Socher R, Radev D.
\newblock SummEval: Re-evaluating Summarization Evaluation.
\newblock Transactions of the Association for Computational Linguistics. 2021
  04;9:391-409.
\newblock Available from: \url{https://doi.org/10.1162/tacl\_a\_00373}.

\bibitem{Chauhan2022ACS}
Chauhan SS, Daniel P.
\newblock A Comprehensive Survey on Various Fully Automatic Machine Translation
  Evaluation Metrics.
\newblock Neural Processing Letters. 2022;55:12663-717.
\newblock Available from:
  \url{https://api.semanticscholar.org/CorpusID:249261826}.

\bibitem{krubinski-etal-2021-just}
Krubi{\'n}ski M, Ghadery E, Moens MF, Pecina P.
\newblock Just Ask! Evaluating Machine Translation by Asking and Answering
  Questions.
\newblock In: Barrault L, Bojar O, Bougares F, Chatterjee R, Costa-jussa MR,
  Federmann C, et~al., editors. Proceedings of the Sixth Conference on Machine
  Translation. Online: Association for Computational Linguistics; 2021. p.
  495-506.
\newblock Available from: \url{https://aclanthology.org/2021.wmt-1.58/}.

\end{thebibliography}

\end{document}